\documentclass[a4paper,UKenglish,cleveref, autoref, thm-restate]{lipics-v2021}

\title{Towards High-Level Modelling in Automated Planning} 

\titlerunning{Towards High-Level Modelling in Automated Planning} 

\author{Carla Davesa}{School of Computer Science, University of St Andrews, UK}{}{}{}

\author{Joan Espasa}{School of Computer Science, University of St Andrews, UK}{}{}{}
\author{Ian Miguel}{School of Computer Science, University of St Andrews, UK}{}{}{}
\author{Mateu Villaret}{Departament d’Informàtica, Matemàtica Aplicada i Estadística, Universitat de Girona, Spain}{}{}{}

\authorrunning{C. Davesa, J. Espasa, I. Miguel and M. Villaret} 

\Copyright{Carla Davesa, Joan Espasa, Ian Miguel and Mateu Villaret} 



\category{} 

\relatedversion{} 

\nolinenumbers 

\EventEditors{Nysret Musliu and Hélène Verhaeghe}
\EventNoEds{2}
\EventLongTitle{23rd Workshop on Constraint Modelling and Reformulation}
\EventShortTitle{ConfWS 2024}
\EventAcronym{}
\EventYear{2024}
\EventDate{September 2––3, 2024}
\EventLocation{Girona, Spain}
\EventLogo{}
\SeriesVolume{}
\ArticleNo{8}

\begin{document}

\maketitle

\begin{abstract}
Planning is a fundamental activity, arising frequently in many contexts, from daily tasks to industrial processes. The planning task consists of selecting a sequence of actions to achieve a specified goal from specified initial conditions. The Planning Domain Definition Language (PDDL) is the leading language used in the field of automated planning to model planning problems. Previous work has highlighted the limitations of PDDL, particularly in terms of its expressivity. Our interest lies in facilitating the handling of complex problems and enhancing the overall capability of automated planning systems. Unified-Planning is a Python library offering high-level API to specify planning problems and to invoke automated planners. In this paper, we present an extension of the UP library aimed at enhancing its expressivity for high-level problem modelling. In particular, we have added an array type, an expression to count booleans, and the allowance for integer parameters in actions. We show how these facilities enable natural high-level models of three classical planning problems.
\end{abstract}

\section{Introduction}
\label{sec:typesetting-summary}

Planning is a fundamental activity, impacting our daily lives in many and varied ways. The planning task consists of selecting a sequence of actions to achieve a specified goal from specified initial conditions. 
This type of problem arises frequently in many contexts, from daily tasks to industrial processes. Examples include scheduling deliveries, organising project schedules, improving manufacturing efficiency, optimising resource allocation, and coordinating transportation routes.
%
Automatically solving planning problems is a central discipline of Artificial Intelligence that involves specifying the desired outcome (the `what') in a purely declarative manner, leaving it to the planning engine to determine the sequence of actions (the `how') needed to reach that outcome.

Consider a scenario where a delivery robot operates within an environment that can be represented as a grid of cells, each represented as distinct locations that the robot can occupy. The robot's objective is to transport a package from one position to another. This involves considering various states such as the robot's presence in a cell, the package's location, and whether the robot is holding the package. The planning task requires defining a set of possible actions: the robot can move between cells, pick up and drop off packages. In the initial state, both the robot and package are located in specific cells, and the goal is to find a sequence of actions where the robot successfully delivers the package to the desired cell.

In such scenarios, it is natural to represent the grid as a matrix. However, traditional planning frameworks do not support these grid-like structures directly, which poses challenges when modelling the problem. This motivates our work to develop new planning constructs and extensions that can effectively model problems in such structured environments.

The difficulty of solving planning problems grows rapidly with their size in terms of the number of states and possible actions considered. Over many years, a great deal of effort by a number of different research groups has resulted in the development of highly efficient AI planners~\cite{PlanningThroughYears}. These planners serve as the computational engines that apply problem-solving algorithms to generate optimal or satisfactory plans given specific problem domains.
However, it is beneficial to consider automated modelling as much a part of the process of solving a planning problem as the search for a solution. The choice of a model has a significant effect on the performance of state-of-the-art AI planning systems~\cite{RepresentationMatters, RepresentationMatters2}, similarly to the importance of modelling to Constraint Programming. Constraint Programming has been successfully used to solve planning problems~\cite{bartak2010constraint, related1} and is particularly well suited to planning problems when the problem requires a certain level of expressivity, such as temporal reasoning or optimality~\cite{vidal2006branching,related2}.

The Planning Domain Definition Language (PDDL)~\cite{PDDL} is the leading language used in the field of automated planning to model planning problems and the domains in which they occur. It provides a formal way to concisely describe the problem in terms of objects, predicates, actions and functions with parameters. PDDL was created in an effort to standardise the input for AI planners, facilitating the solving of planning problems. 

Previous work~\cite{ImpactModellingLanguages} highlighted the
limitations of PDDL, particularly in terms of its expressivity, prompting the development of abstraction techniques to extract higher level concepts from PDDL models. Our interest lies in facilitating the handling of complex problems and enhancing the overall capability of automated planning systems.
The incorporation of high-level concepts, akin to those available in {\sc Essence}~\cite{frisch2008ssence} via {\sc Conjure}~\cite{akgun2022conjure} in Constraint Programming, such as functions, relations, arrays, (multi)sets, and sequences, can significantly enrich the modelling scope and flexibility.

Unified-Planning (UP)~\cite{UP} is a Python library offering high level API to specify planning problems and to invoke planning engines.
This open-source library has gained substantial recognition within the research community due to its extensive adoption and continuous development, making it an excellent candidate on which to base our research efforts, offering ample opportunity for exploration and experimentation.

In this paper, we present an extension of the UP library aimed at enhancing its expressiveness for high-level problem modelling.
The new high-level implementations we are introducing include:
\begin{enumerate}
    \setlength{\itemsep}{1pt} 
    \setlength{\parskip}{0pt} 
    \setlength{\parsep}{0pt} 
    \item A new UP type: Array
    \item A new UP expression: Count
    \item Support for Integers as Parameters in Actions
\end{enumerate}
Furthermore, we have developed three new UP compilers, each dedicated to removing one of the high-level concepts we've implemented. These compilers automate the translation of these constructs into simpler ones, employing advanced problem transformations similar to those used for other features within the library.

In the experimental part of this work, we model three classical planning problems — Plotting~\cite{Plotting}, Rush Hour~\cite{RushHour}, 8-Puzzle~\cite{8puzzle} — to demonstrate the effectiveness of our extended framework. These problems are particularly suitable for array-based modelling due to their grid-based nature, which aligns well with our new array type implementation.

\section{Background and Related Work}
A classical planning problem is typically formalised as a tuple \( \sqcap = \langle F, A, I, G \rangle \), where \( F \) is a set of propositional state variables, \( A \) is a set of actions, \( I \) is the initial state, and \( G \) is the goal.
A \( state \) is a variable-assignment (or valuation) function over state variables \( F \), which maps each variable of \( F \) into a truth value. An action \( a \in A \) is defined as a tuple \( a = \langle \text{Pre}_a, \text{Eff}_a \rangle \), where \( \text{Pre}_a \) refers to the preconditions and \( \text{Eff}_a \) to the effects of the action. Preconditions (\( \text{Pre} \)) and the goal \( G \) are first-order formulas over propositional state variables. Action effects (\( \text{Eff} \)) are sets of assignments to propositional state variables.
An action \( a \) is applicable in a state \( s \) only if its precondition is satisfied in \( s \) (\( s \models \text{Pre}_a \)). The outcome after the application of an action \( a \) will be the state where variables that are assigned in \( \text{Eff}_a \) take their new value, and variables not referenced in \( \text{Eff}_a \) keep their current values. 
A sequence of actions \( \langle a_0, \ldots, a_{n-1} \rangle \) is called a plan. We say that the application of a plan starting from the initial state \( I \) brings the system to a state \( s_n \). If each action is applicable in the state resulting from the application of the previous action and the final state satisfies the goal (i.e., \( s_n \models G \)), the sequence of actions is a \textit{valid plan}. A planning problem has a solution if a valid plan can be found for the problem.

A formalisation for the previously mentioned delivery robot scenario, set within a 2x2 grid with positions named \textit{P00}, \textit{P01}, \textit{P10}, and \textit{P11}, could be as follows:
\begin{align*}
& \sqcap = \langle \\
& \quad F = \{ \text{at\_robot}(p), \text{at\_package}(p), \text{holding\_package} \}, \\
& \quad A = \{ \\
& \quad\quad \text{Move}(p1, p2) = \langle \text{at\_robot}(p1), \{ \text{at\_robot}(p2), \neg \text{at\_robot}(p1)\} \rangle, \\
& \quad\quad \text{PickUp}(p) = \langle \text{at\_robot}(p) \land \text{at\_package}(p), \{ \text{holding\_package}, \neg \text{at\_package}(p) \} \rangle, \\
& \quad\quad \text{DropOff}(p) = \langle \text{at\_robot}(p) \land \text{holding\_package}, \{ \text{at\_package}(p), \neg \text{holding\_package} \} \rangle, \\
& \quad \}, \\
& \quad I = \{ \text{at\_robot}(P00), \text{at\_package}(P10), \neg \text{holding\_package} \}, \\
& \quad G = \{ \text{at\_package}(P11) \} \\
& \rangle
\end{align*}
Is important to note that some of these propositional variables \( F \) are actually first-order atoms over unquantified variables. For instance, the atom \textit{at\_robot(p)} is not a propositional variable by itself. Here, \textit{p} is a parameter that represents a position in the grid and can be substituted with specific grid locations such as \textit{P00, P01, P10 and P11}. During the grounding process, this predicate generates multiple propositional variables corresponding to each specific position in the grid: \textit{at\_robot\_P00, at\_robot\_P01, at\_robot\_P10 and at\_robot\_P11}.
The actions \( A \) define the tasks the robot can perform, such as \textit{Move(p1, p2)}, which allows the robot to move from position \textit{p1} to position \textit{p2} if it is currently in position \textit{p1}. So when the action is executed, the robot moves to position \textit{p2} and it is no longer in position \textit{p1}.
The initial state \( I \) is the in \textit{P00}, the package is in \textit{P10}, and the robot not holding the package.
The goal state \( G \) is to have the package in \textit{P11}.

The Planning Domain Definition Language (PDDL)~\cite{PDDL} initially supported only Boolean types. Over the years, it has evolved significantly to include support for features such as numeric types, temporal constraints, hierarchical types, durative actions, derived predicates and conditional effects. This evolution has greatly enhanced the expressiveness and flexibility of PDDL, allowing for more sophisticated and detailed modelling of planning problems.
Despite these advancements, there remains a significant lack of expressiveness in PDDL, particularly when it comes to representing more complex planning scenarios~\cite{ImpactModellingLanguages}.

The UP library simplifies the processes of both formulating planning problems and utilising automated planners. 
The library allows users to define problems in a simple and intuitive manner and solve them using any of the wide array of supported solvers. Additionally, it provides functionality for exporting and importing problems in PDDL or ANML~\cite{anml} format, and executing advanced problem transformations such as simplification, grounding, and removal of conditional effects.

In the field of automated planning, various approaches have been explored to enhance planning modelling methodologies. This section offers an overview of recent advancements in planning techniques and related works that have inspired or guided our research.

Geffner's Functional STRIPS~\cite{GeffnerFunctionalStrips} introduced an extension of the STRIPS planning language by incorporating first-class function symbols. This addition allowed for greater flexibility in representing planning problems, enabling more efficient encodings and supporting complex tasks with minimal action definitions.
Recent work by Geffner and Frances~\cite{FrancesGeffner} explores how to address the computational challenges posed when solving problems expressed in Functional STRIPS by using a Constraint Satisfaction Problem to compute an heuristic that guides the search.

Planning Modulo Theories (PMT)~\cite{PMT}, inspired by SAT Modulo Theories (SMT), offers a flexible modelling language and framework where arbitrary first-order theories can be treated as parameters. Although further work has been done in the context of PMT~\cite{BofillEV21}, no concrete implementations have been released. 

Elahi and Rintanen~\cite{ComplexDataTypesPDDL} proposed a modelling language supporting complex data types like Booleans, numeric types, enumerated types, records, unions, arrays, sets, and lists, which are reduced to a Boolean representation. This Boolean representation is further reduced to PDDL, allowing existing domain-independent planners to solve problems specified in the richer modelling language. While this approach effectively enhances PDDL's expressivity through the use of complex data types, our work aims to extend these capabilities further by leveraging the UP library. By directly integrating complex data types into UP, we benefit from Python’s simplicity and readability, along with its extensive libraries and community support. This results in intuitive and concise models that are easier to understand and manipulate. Inspired by {\sc Essence} and {\sc Conjure}, our method aims to introduce similar high-level modelling expressivity to automated planning, providing a more intuitive and fluent framework.

\section{Pipeline}

\begin{figure}[t]
    \centering
    \includegraphics[width=0.55\textwidth]{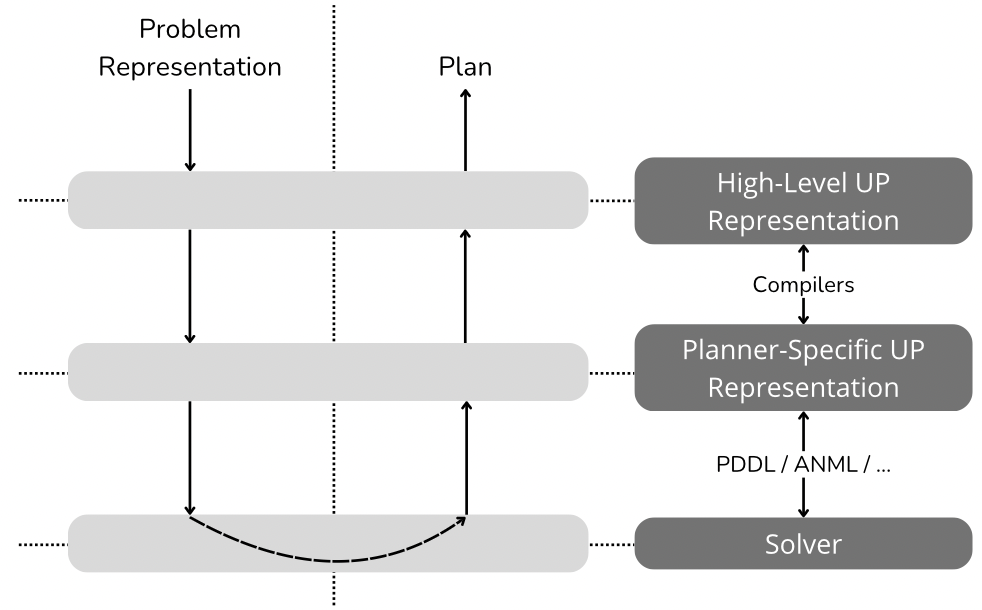}
    \caption{Automated Planning Modelling Pipeline}
    \label{fig:pipeline}
\end{figure}

In this section, we provide an overview of the pipeline within the Unified-Planning framework. The framework is designed to streamline the process of transforming planning problems into formats that various planners can understand and solve. The different stages of this pipeline are depicted in Figure~\ref{fig:pipeline}.

The first step in defining a planning problem is to create a new instance that serves as a container for all the elements that constitute the problem: the fluents, actions, objects, initial state, and goals. 
Similarly to PDDL, objects typically represent entities in the problem domain, each with a type. Note that the term \textit{fluent} has been historically used to refer to state variables that may change over time.
UP also uses a lifted representation of the problem, with state variables and actions having parameters, enabling a concise definition of the problem.

We further distinguish between two representation levels to clearly delimitate when compilers need to be used on elements of the problem that might need to be transformed to ensure compatibility with the planners. It's important to note that if the original problem does not contain any high-level features or the chosen planner fully supports all the problem's features, compilers may not be necessary.

\begin{description}
    \item[High-Level UP Representation] Initially, the planning problem can be defined at a high-level, including complex features such as conditional effects, quantifiers, or user-type fluents. These features might not be supported by all planners, requiring the use of compilers to transform the problem into a compatible `low-level'. We categorise the proposed implementations — array types, count expressions, integer parameters in actions — as high-level representations.
    \item[Planner-Specific UP Representation] After the potential application of compilers, the problem is expressed in a simplified format. This version retains the semantics of the original problem instance while transforming any unsupported features by the targeted planner.
\end{description}

\paragraph*{Specific Planning Languages Representation}
Unified-Planning offers the ability to transform the problem representation into various specific planning languages such as PDDL and ANML.
Whether the problem is translated to PDDL, ANML, or another specific planning language before passing to the solver, depends on the requirements of the chosen planner. Some planners necessitate translating the problem into a specific format before they can process it, essentially when they are designed to exclusively work with these languages.
Alternatively, if the selected planner can directly understand the UP representation of the problem, this can straightforwardly interface with the planner without an additional compilation step.

\paragraph*{Solver}
Finally, the converted planning problem is fed into a planner for execution. UP provides access to a variety of planners, each with distinct capabilities, which can be utilised to solve different types of planning problems.
The planner employs search algorithms and planning techniques to find a plan that transitions from the initial state to the goal state. 

\section{The Proposed UP Extensions}
    \subsection{Array Type}
The new Array Type class is designed to represent arrays consisting of a specified number of elements of a given type.
It relies on two main parameters: \texttt{size} and \texttt{elements\_type}. 
The \texttt{size} parameter signifies the number of elements contained within the array and must be an integer (Python class int) with a predefined value greater than one. 
On the other hand, \texttt{elements\_type} represents the type of the elements within the array. It is optional and it defaults to None, meaning no specific type is assigned. In this case, the array will be assumed to be of Boolean type.
The construction of the class is shown in Listing~\ref{list:array-type}.

\begin{lstlisting}[caption={Construction of the Array Type.} ,label=list:array-type,float=h,abovecaptionskip=-\medskipamount] 
def ArrayType(size: int, elements_type: Type=None) -> unified_planning.model.types.Type
\end{lstlisting}

This type empowers us to represent tables or matrices effectively. With this implementation, we can now define \textit{ArrayType Fluents}, giving us the ability to access the array's elements individually, treating each as a fluent, while also knowing the position of each within the array.
Moreover, given that arrays are considered types themselves and the \texttt{elements\_type} parameter represents a type, we can create arrays of arrays and so forth, enabling the creation of nested arrays.


Assume the initial example of a robot that operates in a 3x3 grid, where the robot can only move to adjacent cells —left, right, up, or down—, but not diagonally. Without using arrays, it is  necessary to define all the relationships between the different cells to determine if they are neighbours. However, by utilising arrays, their position and relationship becomes implicit. For example, we can define the grid cells using a double array-type fluent (matrix) that encapsulates Booleans, indicating whether the robot is in each cell of the grid. The same approach can be used for the package. \Cref{list:array-type-robots} shows the definition of the initial state and fluents. Note that all positions that are not specified in \texttt{initial\_values} default to false.

    \paragraph*{Array Type Compiler}
Given that this new implementation is not compatible with the planners, we've developed an \textit{ArraysRemover} compiler to transform arrays into individual elements.
For each array-type fluent, we create a series of fluents that correspond to its individual elements. These new fluents maintain the original name but are suffixed with '\_' followed by the respective position indices of the array elements. For example, the \(i\)-th element of the array \(my\_ints\), accessed using \(my\_ints[i]\), transforms into a new fluent named \(my\_ints\_i\).
Moreover, this methodology extends to multidimensional arrays. For each dimension, an index is added to represent all the elements. Considering the fluent \texttt{at\_robot} of the previous example, the elements are separated into the nine new fluents with the format depicted in Listing~\ref{list:new-fluents}.
Furthermore, in all actions and goals, we not only replace array accesses, as previously illustrated, but also manage the entire array comprehensively. For instance, when encountering an expression such as \texttt{at\_robot[0] $\iff$ [False, False, False]}, we decompose it into: \texttt{at\_robot\_0\_0} $\iff$ \texttt{False} $\wedge$ \texttt{at\_robot\_0\_1} $\iff$ \texttt{False} $\wedge$ \texttt{at\_robot\_0\_2} $\iff$ \texttt{False}.

\begin{figure}
    \centering
    \begin{minipage}[t]{0.47\textwidth}
        \centering
        \lstset{caption={UP representation of \texttt{at\_robot} fluent using Array-Type.},label=list:array-type-robots,float=h,escapeinside={(*@}{@*)},abovecaptionskip=-\medskipamount}
        \begin{lstlisting}
fluents = [
  array[3, array[3, bool]] at_robot
]
initial fluents default = [
  array[3, array[3, bool]] at_robot := [
    [false, false, false], 
    [false, false, false], 
    [false, false, false]
  ]
]
initial values = [
  at_robot[0][0] := true
]
        \end{lstlisting}
    \end{minipage}%
    \hfill
    \begin{minipage}[t]{0.48\textwidth}
        \centering
        \lstset{caption={UP representation of derived fluents from \texttt{at\_robot} after applying the Array Type Compiler.},label=list:new-fluents,float=h,escapeinside={(*@}{@*)},abovecaptionskip=-\medskipamount}
        \begin{lstlisting}
fluents = [
  bool at_robot_0_0
  bool at_robot_0_1
  bool at_robot_0_2
  ...
  bool at_robot_2_2
]
initial fluents default = [
  bool at_robot_0_0 := false
  bool at_robot_0_1 := false
  bool at_robot_0_2 := false
  ...
  bool at_robot_2_2 := false
]
initial values = [
  at_robot_0_0 := true
]
        \end{lstlisting}
    \end{minipage}
    \label{fig:listings-side-by-side}
\end{figure}

\begin{lstlisting}[caption={UP representation of the \texttt{move\_right} action using Integer-Type Parameters.},label=list:move-right,float=h,abovecaptionskip=-\medskipamount]
action move_right(integer[0, 2] r, integer[0, 1] c) {
  preconditions = [
    at_robot[r][c]
  ]
  effects = [
    at_robot[r][(c + 1)] := true
    at_robot[r][c] := false
  ]
}
\end{lstlisting}

 \paragraph*{Undefinedness}\label{par:undefinedness}

Introducing arrays poses a significant issue when accessing positions outside the arrays' defined range. The concept of undefinedness in planning is not extensively addressed because traditional planning models typically require well-defined states and actions.
In our new implementation, we have been inspired by how undefinedness is handled in constraint programming, based on a study proposing three approaches~\cite{undefinedness}. Two of these methods involve introducing a new truth value, which we find impractical for the library due to substantial modification overhead. The third approach transforms expressions containing undefined values to False. For example, in an array-type fluent \(my\_ints\) containing three integers, accessing \(my\_ints[3]\) (where the array has elements only at indices 0, 1, and 2) would evaluate the expression \((my\_ints[3] == 2)\) as \(False\).

Our solution is a hybrid approach comprising two modes: restrictive and permissive. 
\begin{description} 
    \item[Restrictive] encountering an out-of-range access triggers an error, halting the program with a message indicating the undefined element.
    \item[Permissive] handles out-of-range accesses differently depending on whether they occur in preconditions or effects. In preconditions, akin to constraint handling, if the fluent is a Boolean element, it evaluates itself to False. But, if the fluent is not Boolean, the simplest Boolean expression evaluates to False. In effects, an attempt to access out-of-range simply removes that action, notifying the user via a message.
\end{description}
We illustrate in Listings~\ref{list:ex-undef1} and~\ref{list:ex-undef2} two different examples to demonstrate how the Permissive approach manages various scenarios within the preconditions.

\begin{figure}
    \centering
    \begin{minipage}[t]{0.37\textwidth}
        \centering
        \lstset{caption={UP representation simplifying expressions for Boolean undefined fluents in permissive mode.},label=list:ex-undef1,float=h,escapeinside={(*@}{@*)},abovecaptionskip=-\medskipamount}
        \begin{lstlisting}
And(at_robot[2][2],(*@\textcolor{red}{at\_robot[2][3]}@*))
And(at_robot[2][2],(*@\textcolor{red}{False}@*))
(*@\textcolor{red}{False}@*)
        \end{lstlisting}
    \end{minipage}%
    \hfill
    \begin{minipage}[t]{0.6\textwidth}
        \centering
        \lstset{caption={UP representation simplifying expressions for non-Boolean undefined fluents in permissive mode.},label=list:ex-undef2,float=h,escapeinside={(*@}{@*)},abovecaptionskip=-\medskipamount}
        \begin{lstlisting}
Or(Equals(Plus((*@\textcolor{red}{my\_ints[3]}@*),2),2),Equals(my_ints[2],0))
Or((*@\textcolor{red}{False}@*),Equals(my_ints[2],0)
Equals(my_ints[2],0)
        \end{lstlisting}
    \end{minipage}
    \label{fig:listings-side-by-side}
\end{figure}

\subsection{Integer Type Parameters in Actions}
\label{sec:int-param-act}

We identified support for bounded integer parameters to be passed to actions as a highly useful implementation when working with arrays.

We will illustrate this approach with the previous example. 
To properly control the possible movements, we create four actions, each corresponding to one direction. For instance, with the \texttt{move\_right} action, shown in Listing~\ref{list:move-right}, we ensure that the robot cannot move outside the grid by allowing this movement only when the robot is in one of the first two columns. This means that only the integers \(0\) and \(1\) will be passed as parameters for the columns, preventing the robot from moving right when it is in the rightmost column. 

This approach allows for more flexible and concise problem descriptions in planning scenarios. It enables us to refer to any element within the array by indexing this integer in the array, eliminating the need to create an action for each individual element.
However, it is crucial to ensure that the bounded integer values fall within the specified domain range, as values outside this range may lead to undefined behaviour. It is crucial to understand how undefined values will be handled according to the strategies outlined above, ensuring the implementation appropriately manages any out-of-range values introduced.
Additionally, integers can be utilized not only for indexing arrays but also as values in preconditions or effects, as well as in arithmetic operations, enhancing the model's flexibility and expressiveness.

    \paragraph*{Integer Type Parameters in Actions Compiler}

This compiler overcomes the limitations of the previous implementation, as planners could not comprehend or process integer parameters or anything other than objects in actions.
For each action, the compiler generates new actions for each possible combination of the integer parameters. These new actions do not include the integer parameters, as they are replaced by their respective values within the preconditions and effects. Retaining the original name, they are suffixed with `\_' followed by the current integer parameter values separated by `\_'.

For instance, when the compiler processes the \texttt{move\_right} action described in Listing~\ref{list:move-right}, it will transform it into six new actions, one for each possible combination of the parameters \texttt{r} and \texttt{c} — \((0,0), (0,1), (1,0), (1,1), (2,0), (2,1)\) —. 
In Listing~\ref{list:new-actions}, the action generated for the parameters (0,1) illustrates how the parameters \texttt{r} and \texttt{c} in both preconditions and effects are substituted with their specific values.
Furthermore, this compiler simplifies expressions wherever possible during parameter substitution with integer values. This simplifier handles arithmetic operations at runtime, reducing the size and complexity of the generated expressions. For example, in the previous instance with \texttt{at\_robot[0][2]}, the value \(2\) results from the addition (\(1+1\)), which is computed automatically during compilation. This optimisation streamlines execution, enhances efficiency, and simplifies the problem description for the planner. 

\begin{figure}
    \centering
    \begin{minipage}[t]{0.33\textwidth}
        \centering
        \lstset{caption={UP representation of \texttt{move\_right(r=0, c=1)} action after applying the Integer-Type Parameters Compiler.},label=list:new-actions,float=h,escapeinside={(*@}{@*)},abovecaptionskip=-\medskipamount}
        \begin{lstlisting}
action move_right_0_1 {
  preconditions = [
    at_robot[0][1]
  ]
  effects = [
    at_robot[0][2] := true
    at_robot[0][1] := false
  ]
}
        \end{lstlisting}
    \end{minipage}%
    \hfill
    \begin{minipage}[t]{0.65\textwidth}
        \centering
        \lstset{caption={Construction of the Count Expression.},label=list:count,float=h,escapeinside={(*@}{@*)},abovecaptionskip=-\medskipamount}
        \begin{lstlisting}
def Count(self, 
  *args: Union[BoolExpression, Iterable[BoolExpression]]
) -> unified_planning.model.fnode.FNode
        \end{lstlisting}
        \lstset{caption={UP representation of a goal ensuring only one room is occupied by robots using the Count Expression.},label=list:count-example,float=h,escapeinside={(*@}{@*)},abovecaptionskip=-\medskipamount}
        \begin{lstlisting}
goals = [
  (Count(at_robot[0][0], at_robot[0][1], at_robot[0][2], 
    at_robot[1][0], at_robot[1][1], at_robot[1][2], 
    at_robot[2][0], at_robot[2][1], at_robot[2][2]) == 1)
]
        \end{lstlisting}
    \end{minipage}
\end{figure}

\subsection{Count Expression}

When encoding problems, we encounter situations where evaluating the number of True statements among multiple Boolean expressions becomes necessary. To address this, we have developed the \texttt{Count} expression, specifically designed to manage Boolean n-arguments efficiently. The purpose of this function is to return an integer representing the number of True expressions among multiple Boolean expressions.
The construction of this expression is shown in Listing~\ref{list:count}, and can be formulated in two ways: \(Count(a,b,c)\) or \(Count([a,b,c])\), where \(a,b,c\) represent Boolean arguments.

We can exemplify the function with an illustration of a possible goal from the previous example. Suppose our problem involves multiple robots, and the goal is for all robots to end up in the same room, leaving only one room occupied. To achieve this, we need to count how many cells are occupied by any robot. This is demonstrated in Listing~\ref{list:count-example}.

    \paragraph*{Count Expression Compiler}
This compiler translates each \textit{Count} expression into a new set of expressions that the planner can comprehend, involving the creation of a new function (known as an integer fluent in Unified-Planning) for each argument of every \textit{Count} expression in the problem.
These functions are designed to represent the Boolean value of each argument: they take on a value of 0 if the expression is False and 1 if it is True. The new fluents will be named sequentially as \(count\_0, count\_1, count\_2\), and so on, for each argument of the expressions in the problem. When different \textit{Count} expressions share an argument with the same expression, they will be assigned the same name.
Moreover, substitutes each argument in the expression with its corresponding function, and replaces the \textit{Count} operator expression with the well-known \textit{Plus} operator, which adds up various integer expressions. This way, we sum up the new functions created, each corresponding to its original Boolean expression. This is shown in Listing~\ref{list:count-remover}.
These functions are initialised depending on the initial value of the fluents and the evaluation of the expression. 
As depicted in Listing~\ref{list:count-initial-values}, in our example, only the position \((0,0)\) is set to True, resulting in the function related to this Boolean expression, \texttt{count\_0}, having an initial value of 1.

\begin{figure}
    \centering
    \begin{minipage}[t]{0.62\textwidth}
        \centering
        \lstset{caption={UP representation of the previous goal after applying the Count compiler.},label=list:count-remover,float=h,escapeinside={(*@}{@*)},abovecaptionskip=-\medskipamount}
        \begin{lstlisting}
goals = [
  (Plus(count_0, count_1, count_2, count_3, count_4, 
        count_5, count_6, count_7, count_8) == 1)
]
        \end{lstlisting}
        \lstset{caption={UP representation of \texttt{move\_right(r=0, c=0)} action effects after applying the Count compiler.},label=list:count-new-action,float=h,escapeinside={(*@}{@*)},abovecaptionskip=-\medskipamount}
        \begin{lstlisting}
action move_right_0_0 {
  preconditions = [
      at_robot_0_0
  ]
  effects = [
    at_robot_0_1 := true
    at_robot_0_0 := false
    count_0 := 0
    count_1 := 1
  ]
}
        \end{lstlisting}
    \end{minipage}%
    \hfill
    \begin{minipage}[t]{0.33\textwidth}
        \centering
        \lstset{caption={UP representation of \texttt{at\_robot} initial values after applying the Count compiler.},label=list:count-initial-values,float=h,escapeinside={(*@}{@*)},abovecaptionskip=-\medskipamount}
        \begin{lstlisting}
initial values = [
  at_robot_0_0 := true
  at_robot_0_1 := false
  ...
  at_robot_2_2 := false
  count_0 := 1
  count_1 := 0
  count_2 := 0
  count_3 := 0
  count_4 := 0
  count_5 := 0
  count_6 := 0
  count_7 := 0
  count_8 := 0
]
        \end{lstlisting}
    \end{minipage}
    \label{fig:listings-side-by-side}
\end{figure}

Furthermore, for each action, if any fluent is modified, those expressions (arguments from Count expressions) containing that fluent will also be evaluated to potentially change the value of the corresponding function based on the evaluation result. As illustrated in Listing~\ref{list:count-remover}, the Integer Type Parameters in Actions compiler and the Array Type Compiler have already been applied. It must be in this order, as discussed in Section~\ref{sec:seq-compilers}. The action \texttt{move\_right\_0\_0} modifies both \texttt{at\_robot\_0\_0} and \texttt{at\_robot\_0\_1}. Consequently, the corresponding functions, \texttt{count\_0} and \texttt{count\_1}, are re-evaluated to their new values: 0 if the Boolean evaluates to False and 1 if it evaluates to True. If the effect is conditional, the function's effect will adhere to the same condition. Moreover, if the effect's fluent has parameters, a condition will be added in this new effect, checking whether these parameters match those of the same fluent in the \textit{Count} argument.

\subsection{Sequence of Compiler Application}
\label{sec:seq-compilers}

To ensure the proper utilisation of the compilers we've developed, it is crucial to follow a specific sequence in their application.
The required order, assuming all three implementations are included in a problem, is as follows:

\begin{enumerate}
    \item Integer Parameters in Actions Compiler
    \item Array Type Compiler
    \item Count Expression Compiler
\end{enumerate}

When a problem includes integer parameters in actions it is crucial to apply the \textit{Integer Parameters in Actions Compiler} before any other compiler. Not applying this compiler first could lead to issues if our problem also involves arrays. For example, accessing specific elements like \texttt{my\_ints[i]} where \texttt{i} is an integer parameter requires knowing its value to determine the array element. Applying the \textit{Array Type Compiler} first without this information would result in errors due to the undefined value of \texttt{i}.

The \textit{Count Expression Compiler} should be applied after resolving complexities related to arrays using the other two compilers. This sequence is critical because the expressions it handles may reference array elements.
Additionally, to effectively adjust the new functions associated with each argument of count expressions, it is essential to first evaluate the changes affecting the fluents within the action effects. Without substituting integers and removing arrays beforehand, we would not know which fluents are affected by action changes. Consequently, we would be unable to implement the required adjustments to our functions when evaluating expressions with the updated values.

\section{Experiments}
In this section, we evaluate our new UP-extended implementation encoding three typical planning games: Plotting, Rush-Hour, and 8-Puzzle, against existing PDDL models. 
These experiments were conducted using a cluster comprising 20 nodes, each equipped  Intel(R) Xeon(R) E-2234 CPUs @ 3.60 GHz with 16GB of RAM.
The solvers used in our experiments are \emph{Enhsp-opt}~\cite{enhsp} for our UP-extended model and \emph{Fast-Downward}~\cite{downward} with the \emph{seq-opt-lmcut} heuristic for the selected PDDL models. 
We employ two different planners to leverage their respective strengths: \emph{Enhsp} (version 20) supports numeric extensions crucial for our model, while \emph{Fast-Downward} (version 23.06+), although lacking numeric support, provides robust heuristic-based search capabilities.
Our aim was to measure the combined preprocessing and solving time, with a configured timeout of 1 hour (all results are presented in seconds), and see if the proposed UP extensions had a reasonable cost regarding solving time. 
For each problem we will highlight the advantages in modelling provided by some of the features that we have developed in UP.
The models with UP-extended, along with the implementation of the proposed extensions, can be found on GitHub at \textit{https://github.com/stacs-cp/unified-planning/tree/new\_types2}.

\begin{figure}[!htb]
    \centering
    \begin{subfigure}{0.34\textwidth}
        \centering
        \includegraphics[width=\textwidth]{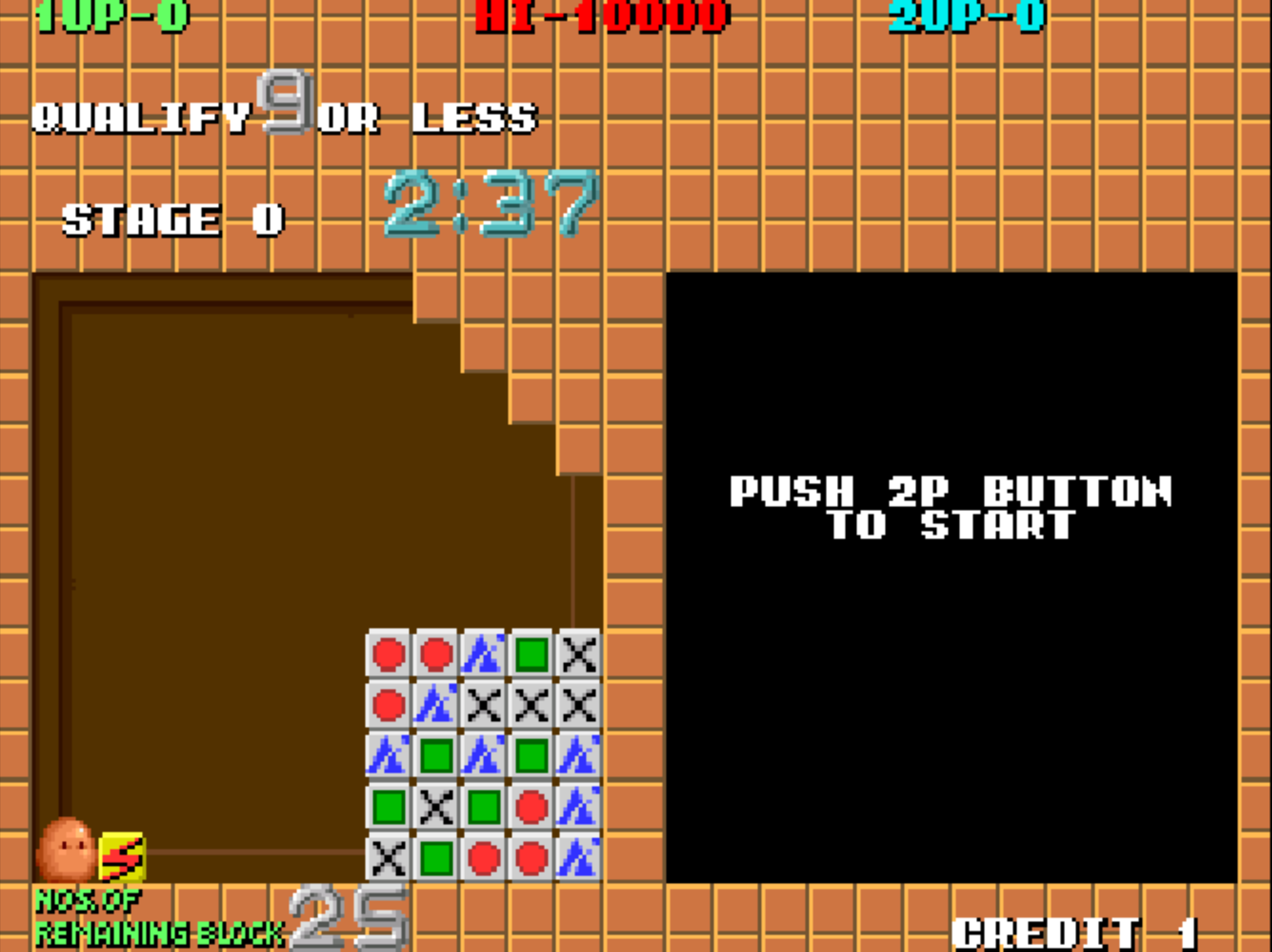}
        \caption{Example of a 5x5 grid Initial Plotting Instance}
        \label{fig:plotting}
    \end{subfigure}
    \hspace{0.05\textwidth}
    \begin{subfigure}{0.23\textwidth}
        \centering
        \includegraphics[width=\textwidth]{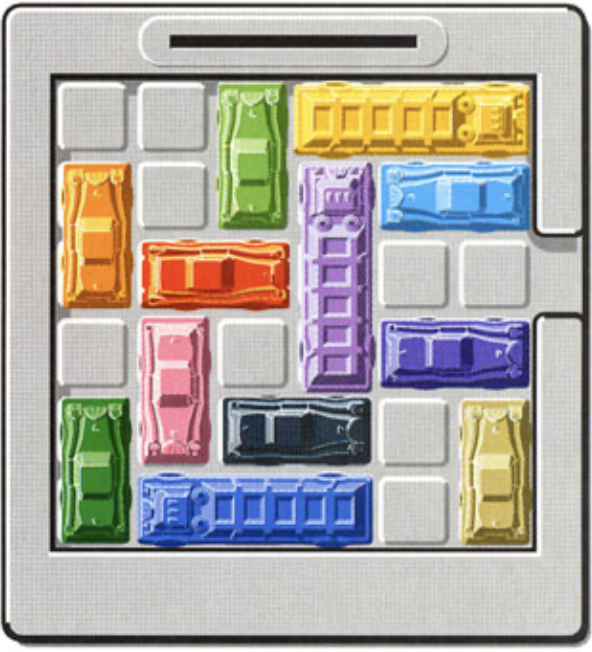}
        \caption{Example of an Initial Rush Hour Instance}
        \label{fig:rushhour}
    \end{subfigure}
    \hspace{0.05\textwidth}
    \begin{subfigure}{0.26\textwidth}
        \centering
        \includegraphics[width=\textwidth]{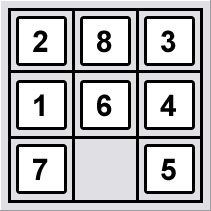}
        \caption{Example of an Initial 8-Puzzle Instance}
        \label{fig:8-puzzle}
    \end{subfigure}
    \caption{Examples of different instances of Plotting, Rush-Hour and 8-Puzzle games.}
    \label{fig:examples}
\end{figure}

    \subsection{Plotting}
Plotting is a tile-matching puzzle video game published by Taito (Figure~\ref{fig:plotting} illustrates an instance of the game). The objective of the game is to remove at least a certain number of coloured blocks from a grid by sequentially shooting blocks into the same grid. The interest and difficulty of Plotting is due to the complex transitions after every shot: various blocks are affected directly, while others can be indirectly affected by gravity.

Listing~\ref{list:plot-model} depicts the grid represented as a double array, with each cell indicating the color of a block. The initial grid configuration is defined using a Python nested list.

\begin{lstlisting}[caption={UP extended: Defining the Plotting game problem instance using Array Type.},label=list:plot-model,float=h,abovecaptionskip=-\medskipamount]
grid = [[R,R,B,G,Y],[R,B,Y,Y,Y],[B,G,B,G,B],[G,Y,G,R,B],[Y,G,R,R,B]]
blocks = Fluent(`blocks', ArrayType(rows, ArrayType(columns, Colour)))
plotting_problem.add_fluent(blocks)
plotting_problem.set_initial_value(blocks, grid)
\end{lstlisting}

We define several actions that manipulate the grid of colored blocks based on the shot. These actions utilise integer parameters to reference different blocks, and we ensure these parameters are within valid ranges to prevent out-of-bounds access.
One such action, detailed in Listing~\ref{list:def-sfr}, is \texttt{shoot\_partial\_row}, which clears blocks of a colour \texttt{p} from a  row \texttt{r} up to the last column \texttt{l} of that row, stopping when the next block is of a different color.

\begin{lstlisting}[caption={UP extended: Defining the Shoot-Full-Row action.},label=list:def-sfr,float=h,abovecaptionskip=-\medskipamount]
spr = unified_planning.model.InstantaneousAction(`shoot_partial_row', p=Colour, 
    r=IntType(0, rows-1), l=IntType(0, columns-2))
spr.add_precondition(Not(Or(Equals(p, W), Equals(p, N))))
for c in range(0, columns-1):
  spr.add_precondition(Or(GT(c,l), Equals(blocks[r][c], p), Equals(blocks[r][c], N)))
spr.add_precondition(Or(
    *[And(Equals(blocks[r][c], p), LE(c,l)) for c in range(columns-1)]
))
spr.add_effect(hand, blocks[r][l+1])
spr.add_effect(blocks[r][l+1], p)
for c in range(0, columns-1):
  spr.add_effect(blocks[0][c], N, LE(c,l))
  for a in range(1, rows):
    spr.add_effect(blocks[a][c], blocks[a-1][c], And(LE(c,l), LE(a,r)))
\end{lstlisting}

In Listing~\ref{list:def-sfr}, we highlight some of the most interesting preconditions and effects of this action, particularly to demonstrate the effectiveness of Python functions, such as \textit{for loops}, for iterating over different elements.
The first precondition ensures that the next block of the column \texttt{l} is different from \texttt{p} and not \texttt{N}, where \texttt{N} indicates no block is present.
The first loop ensures that all elements in the row until the column \textit{l} are either \texttt{p} or \texttt{N}. And the following precondition confirms that at least one of these elements (i.e., among all elements being cleared) is \texttt{p}, preventing unnecessary actions when all blocks are \texttt{N}.
The effects described detail how the action modifies the grid. Initially, the \texttt{hand} takes on the value of the next block in sequence (the first block different), while this position saves the value of \texttt{p}.
Additionally, the nested loops update the grid's elements in accordance with gravity, ensuring the correct movement of blocks above the affected row.

As shown in Listing~\ref{list:def-goal-plotting}, the objective of the game is to have no more than a specified number of blocks remaining, in this example, 4 or fewer. Our new implementation of the \textit{Count} expression is particularly useful for this purpose as it facilitates the counting of blocks that are different from \texttt{N}, indicating how many blocks are left on the grid.

\begin{lstlisting}[caption={UP extended: Defining the Goal.},label=list:def-goal-plotting,float=h,abovecaptionskip=-\medskipamount]
remaining = [Not(Equals(blocks[i][j], N)) for i in range(rows) for j in range(columns)]
plotting_problem.add_goal(LE(Count(remaining), 4))
\end{lstlisting}

\begin{figure}
    \centering
    \includegraphics[width=1\textwidth]{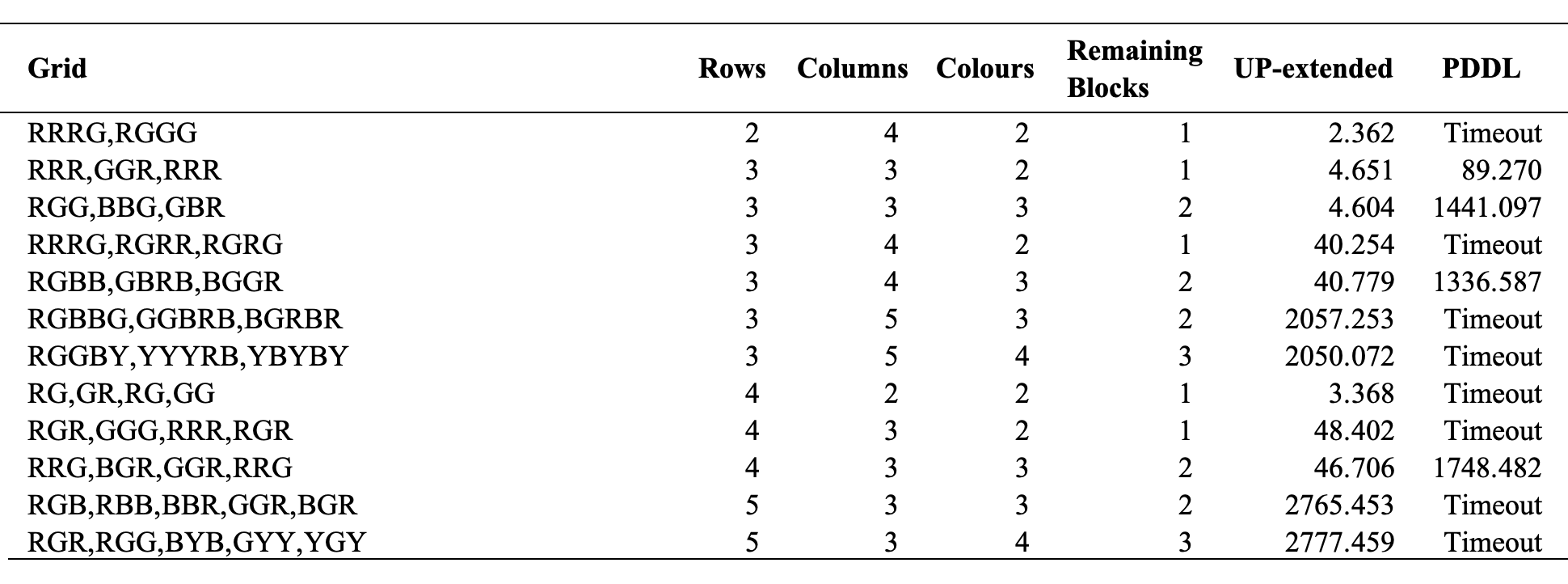}
    \caption{Comparison of Plotting models: UP-extended vs PDDL.}
    \label{fig:plotting-results}
\end{figure}

For the experimental part, we compared our UP-extended model with the PDDL extracted from the paper `Challenges in Modelling and Solving Plotting with PDDL' by J. Espasa et al.~\cite{Plotting2}. We utilised the instances from the database of the same work, which consists of 522 instances.
In Figure~\ref{fig:plotting-results}, we show results for several instances based on grid size, number of colours, and number of remaining blocks. Our model successfully solves 216 instances out of the total, whereas the PDDL model solves only 78 instances.
Our high-level implementations has resulted in a more concise, understandable, and manageable model in comparison to the one of~\cite{Plotting2} where the authors needed to simulate cell positions by ad-hoc encoding numbers and relational predicates (as $\leq$) into PDDL.

    \subsection{Rush Hour}
Rush Hour is a sliding block puzzle game set on a 6x6 grid, where blocks represent vehicles stuck in a traffic jam (Figure~\ref{fig:rushhour} illustrates an instance of the game). The objective is to move a special vehicle, the red car, to the exit located at the right edge of the grid. However, the movement of vehicles is restricted: they can only move forwards or backwards in a straight line and cannot cross over each other.

In modelling the Rush Hour game, we utilise a 6x6 grid represented by a double array, where each cell denotes a letter representing a vehicle, as depicted in Listing~\ref{list:rush-hour-model}.
This grid instantiation process utilises Michael Fogleman's database~\cite{RushHour}, which provides string representations where `o' denotes an empty cell, and each letter represents a distinct vehicle. Vehicles include cars, which occupy 2 cells, and trucks, which occupy 3 cells.
Python's ability to manage strings as iterable sequences allows for a concise representation of the grid and enhances flexibility in modifying its configuration.

\begin{lstlisting}[caption={Defining the Rush Hour game problem instance using Array Type.},label=list:rush-hour-model,float=h,abovecaptionskip=-\medskipamount]
occupied = Fluent('occupied', ArrayType(6, ArrayType(6, Vehicle)))
rush_hour_problem.add_fluent(occupied)
grid = `GBBoLoGHIoLMGHIAAMCCCKoMooJKDDEEJFFo'
for i, char in enumerate(grid):
    r, c = divmod(i, columns)
    if char == '.':
        rush_hour_problem.set_initial_value(occupied[r][c], none)
    else:
        obj = Object(f'{char}', Vehicle)
        if not rush_hour_problem.has_object(char):
            rush_hour_problem.add_object(obj)
            rush_hour_problem.set_initial_value(is_car(obj), grid.count(char) == 2)
        rush_hour_problem.set_initial_value(occupied[r][c], obj)
\end{lstlisting}

\begin{lstlisting}[caption={Defining the Move-Horizontal-Car action.},label=list:rush-hour-mhc,float=h,abovecaptionskip=-\medskipamount]
mhc = unified_planning.model.InstantaneousAction(`move_horizontal_car', v=Vehicle, 
    r=IntType(0,rows-1), c=IntType(0,columns-2), m=IntType(-(columns-2), columns-2))
\end{lstlisting}

Utilising the permissive mode when handling undefinedness (See \nameref{par:undefinedness} above) is highly beneficial in encoding the Rush Hour problem. It eliminates the requirement to specify exact movement possibilities for each vehicle position while ensuring they remain within the grid. Depending on the vehicle's location, it may have a varying number of possible moves without exiting the grid. 
By using this mode, we can define the range of movements from a minimum of \texttt{1} up to the maximum possible movement. For instance, in the \texttt{move\_horizontal\_car} action, illustrated in Listing~\ref{list:rush-hour-mhc}, this maximum distance is \texttt{columns-2} when the car starts at one edge and moves towards the opposite edge.
The permissive mode automatically filters out actions that exceed defined movement ranges, ensuring that actions are generated only for scenarios where vehicles can move within specified cell limits without leaving the grid. 
This optimization streamlines action creation by defining movement ranges at a higher level, eliminating the need to specify possible moves for each individual cell.

\begin{figure}
    \centering
    \includegraphics[width=1\textwidth]{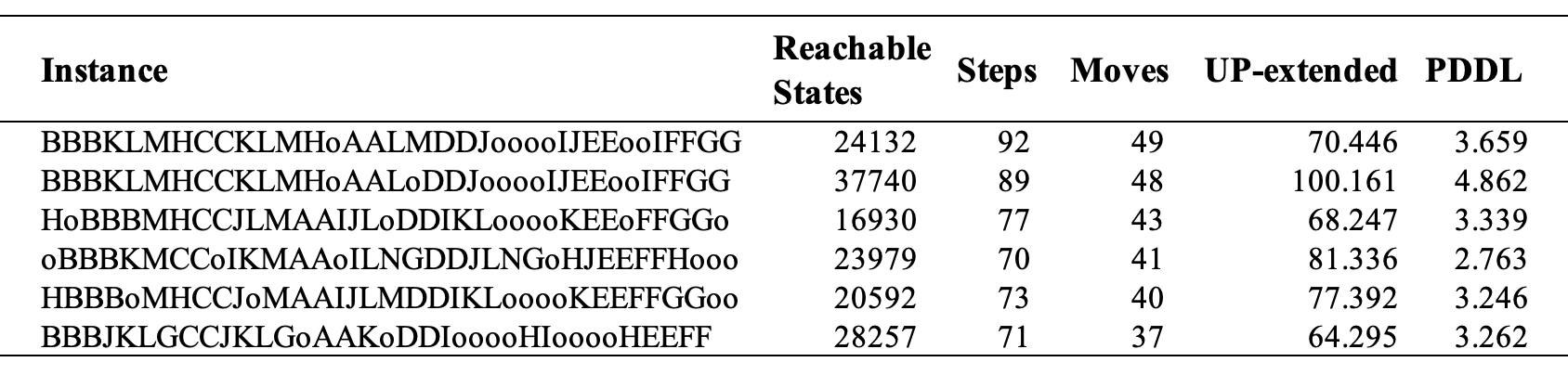}
    \caption{Comparison of Rush Hour models: UP-extended vs PDDL}
    \label{fig:rush-hour-results}
\end{figure}

We compare our UP-extended model with an adapted version of ehajdin's PDDL model, available on GitHub~\cite{modelpddlrush}. Originally designed to restrict vehicle movements to one step per move, we modified it to enable full vehicle mobility, allowing movements ranging from 1 to 4 steps per move.
From Fogleman's database, we selected the 43 most complex instances based on factors influencing difficulty outlined in the undergraduate thesis~\cite{davesa2023rush}. These instances are used to evaluate and compare the effectiveness of both models.
From this selection, we highlight the 6 most difficult instances in Figure~\ref{fig:rush-hour-results}.
While the model does not show improved efficiency compared to the PDDL implementation, it notably enhances problem description fluency and clarity, and significantly reduces model size.

    \subsection{8-Puzzle}
The N-puzzle is a classic sliding puzzle game consisting of a \(k * k\) grid with \(((k * k) - 1)\) numbered tiles (N) and one blank space (Figure~\ref{fig:8-puzzle} illustrates an instance of the game). The objective is to rearrange the tiles by sliding them horizontally or vertically into the blank space, with the goal of achieving a specific configuration, often the ordered sequence of numbers from 1 to N.

\begin{lstlisting}[caption={Defining the 8-Puzzle grid.},label=list:npuzzle-grid,float=h,abovecaptionskip=-\medskipamount]
puzzle = Fluent('puzzle', ArrayType(k, ArrayType(k, IntType(0,8))))
\end{lstlisting}

The definition of the grid, depicted in Listing~\ref{list:npuzzle-grid}, is represented as a 2D array where each cell contains an integer value ranging from 0 to n. The value 0 represents the empty space.

The representation of \texttt{slide\_up} action is shown in Listing~\ref{list:slide-up}, which moves a tile up into the empty space in the grid. The parameters \texttt{r} and \texttt{c} are integers representing the row and column indices of the grid. Note that the range for \texttt{r} is from \(1\) to \(k-1\), since you cannot slide up from the first row.
The first precondition ensures that the tile above the current position (\texttt{r}, \texttt{c}) is the empty space, making the slide-up move possible. The effects describe the result of the action: the tile at the current position moves up and original position is now empty.

\begin{lstlisting}[caption={Defining the Slide-Up action.},label=list:slide-up,float=h,abovecaptionskip=-\medskipamount]
su = unified_planning.model.InstantaneousAction('slide_up', r=IntType(1,k-1), 
    c=IntType(0,k-1))
su.add_precondition(Equals(puzzle[r-1][c], 0))
su.add_effect(puzzle[r-1][c], puzzle[r][c])
su.add_effect(puzzle[r][c], 0)
\end{lstlisting}

To evaluate the performance of our model, we selected the two instances requiring the highest number of steps for a solution, both needing 31 steps, and the two configurations with the highest number of solutions, each having 64 solutions and requiring 30 steps, as detailed in the paper~\cite{8puzzle}. We obtained the PDDL model for these instances from the GitHub repository of the user mazina~\cite{modelpddl8puzzle}.

\begin{figure}
    \centering
    \includegraphics[width=0.65
    \textwidth]{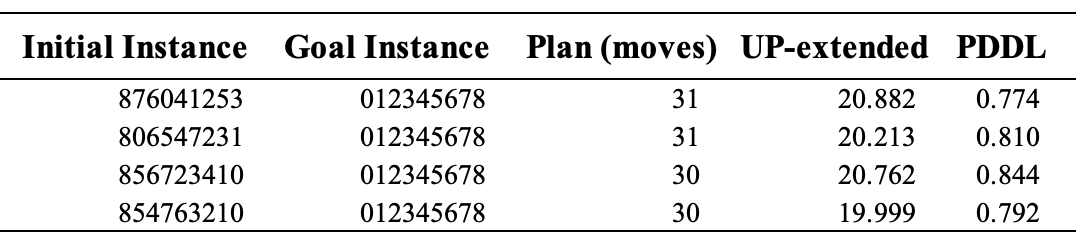}
    \caption{Comparison of 8-puzzle models: UP-extended vs PDDL}
    \label{fig:8-puzzle-results}
\end{figure}

We observe the results in Figure~\ref{fig:8-puzzle-results}. 
As with the Rush Hour model, our implementation did not outperform the PDDL implementation in terms of solving time. However, it also significantly enhanced the clarity and natural representation of the problem, while also reducing the model size.

\section{Conclusions and Future Work}
This proposed extension allows plans to be expressed more naturally, facilitating the management of complex problems. By making the modelling process more intuitive, it significantly reduces the manual effort and time required to select an optimal modelling approach.
One significant advantage of this Python-based framework is its ability to effectively utilise Python functions, such as \textit{for loops}, for iterating over different elements, allowing better manipulation of complex structures. This capability enhances the flexibility and power of the modelling process, making it easier to handle intricate scenarios.
A key benefit of this pipeline is the significant modelling flexibility it provides, enabling the use of different compilers depending on the planner, which illustrates the diverse methods to model, transform and solve the same problem. 
As we have seen, arrays have been particularly useful in our examples. 

Looking ahead, we aim to implement more high-level concepts such as functions, relations, (multi)sets, and sequences to further enhance the modelling capabilities of the framework. We also want to explore alternative compilers for each feature obtaining different possible encoding paths from our high-level UP extensions to PDDL (as done in  constraint programming  with Conjure~\cite{akgun2022conjure} between Essence and Essence Prime).
We also plan to conduct an experimental analysis of the solving time cost of using our proposed high-level UP representations.

\bibliography{lipics-v2021-sample-article}

\appendix

\end{document}